\documentclass[11pt,a4paper]{article}
\usepackage{authblk}

\usepackage[hyperref]{acl2019}
\usepackage{times}
\usepackage{latexsym}
\usepackage{graphicx}
\usepackage{url}
\usepackage{todonotes}
\usepackage[utf8]{inputenc}
\usepackage[english]{babel}
\usepackage[T1]{fontenc}
\usepackage{csquotes}
\usepackage{pgf}
\usepackage{tikz}
\usepackage{booktabs}
\usepackage{xcolor}
\usepackage{amsmath}
\usepackage{comment}
\usepackage{enumitem}
\usepackage{paralist} 

\usepackage[normalem]{ulem}
\usepackage[ruled]{algorithm2e}
\usepackage{tikz}
\usetikzlibrary{fit,positioning, calc}

\tikzset{
    between/.style args={#1 and #2}{
         at = ($(#1)!0.5!(#2)$)
    }
}

\title{From the Paft to the Fiiture: a Fully Automatic NMT and Word Embeddings Method for OCR Post-Correction}

\author{Mika H\"am\"al\"ainen$^{\spadesuit}$~~~Simon Hengchen$^{\diamondsuit}$\\

$^{\spadesuit}$ Department of Digital Humanities, University of Helsinki\\
$^{\diamondsuit}$ COMHIS, University of Helsinki \\
{\tt firstname.lastname@helsinki.fi}}

\date{April 2019}

\aclfinalcopy

\begin{document}

\maketitle

\begin{abstract}
A great deal of historical corpora suffer from errors introduced by the OCR (optical character recognition) methods used in the digitization process. Correcting these errors manually is a time-consuming process and a great part of the automatic approaches have been relying on rules or supervised machine learning. We present a fully automatic unsupervised way of extracting parallel data for training a character-based sequence-to-sequence NMT (neural machine translation) model to conduct OCR error correction.
\end{abstract}

\section{Introduction}

Historical corpora are a key resource to study social phenomena such as language change in a diachronic perspective. Approaching this from a computational point of view is especially challenging as historical data tends to be noisy. The noise can come from OCR (optical character recognition) errors, or from the fact that the spelling conventions have changed as the time has passed, as thoroughly described by \citet{piotrowski2012natural}.

However, depending on the NLP or DH task being modelled, some methods can cope with the noise in the data. 
Indeed, \citet{hill-hengchen2019OCR} use a subset of an 18th-century corpus, ECCO,\footnote{Eighteenth Century Collections Online (ECCO) is a dataset which \enquote{contains over 180,000 titles (200,000 volumes) and more than 32 million pages}, according to its copyright holder Gale: \url{https://www.gale.com/primary-sources/eighteenth-century-collections-online}.} and its ground truth version, ECCO-TCP,\footnote{ECCO-TCP (Text Creation Partnership) \enquote{is a keyed subset of ECCO,
compiled with the support of 35 libraries and made up of 2,231 documents}. \citep{hill-hengchen2019OCR}} to compare the output of different common DH methods such as authorship attribution, count-based vector space models, and topic modelling, and report that those analyses produce statistically similar output despite noisiness due to OCR. Their conclusion is similar to \citet{rodriquez2012comparison} in the case of NER and to \citet{franzini2018attributing} in the case of authorship attribution, but different from \citet{mutuvi2018evaluating} who, specifically on topic modelling for historical newspapers, confirm the often repeated trope of data too dirty to use.
However, reducing the noise of OCRed text by applying a post-correction method makes it possible to gain the full potential of the data without having to re-OCR it and opens up the possibility to process it with the myriad of more precise NLP tools designed for OCR-error free text.

This paper focuses on correcting the OCR errors in ECCO. We present an unsupervised method based on the advances neural machine translation (NMT) in historical text normalization\footnote{Our code https://github.com/mikahama/natas}. As NMT requires a parallel dataset of OCR errors and their corresponding correct spellings, we propose a method based on word embeddings, a lemma list, and a modern lemmatizer to automatically extract parallel data for training the NMT model.

\section{Related Work}
\label{sec:relatedwork}
OCR quality for historical texts has recently received a lot of attention from funding bodies and data providers. 
Indeed, \citet{smith2019OCR} present a (USA-focused) technical report on OCR quality, and aim to spearhead the efforts on setting a research agenda for tackling OCR problems.
Other initiatives such as \citet{adesam2019exploring} set out to analyse the quality of OCR produced by the Swedish language bank Spr{\aa}kbanken, \citet{drobac2017ocr} correct the OCR of Finnish newspapers using weighted finite-state methods, \citet{tanner2009measuring} measure mass digitisation in the context of British newspaper archives, while the European Commission-funded IMPACT project\footnote{\url{http://www.impact-project.eu}} gathers 26 national libraries and commercial providers to \enquote{take away the barriers that stand in the way of the mass digitization of the European cultural heritage} by improving OCR technology and advocating for best practices.

\citet{dong2018multi} present an unsupervised method for OCR post-correction. As opposed to our character-level approach, they use a word-level sequence-to-sequence approach. As such a model requires training data, they gather the data automatically by using repeated texts. This means aligning the OCRed text automatically with matched variants of the same text from other corpora or within the OCRed text itself. In contrast, our unsupervised approach does not require any repetition of text, but rather repetition of individual words.

Different machine translation approaches have been used in the past to solve the similar problem of text normalization, which means converting text written in a non-standard form of a language to the standard form in order to facilitate its processing with existing NLP tools. SMT (statistical machine translation) has been used previously, for instance, to normalize historical text \cite{pettersson2013smt} to modern language and to normalize modern Swiss German dialects \cite{swissgerman} into a unified language form. More recently with the rise of the NMT, research has emerged in using NMT to normalize non-standard text, for example work on normalization of medieval German \cite{korchagina2017normalizing} and on historical English \cite{ceec_normal_santafe}.

All of the normalization work cited above on using machine translation for normalization has been based on character-level machine translation. This means that words are split into characters and the translation model will learn to translate from character to character instead of word to word. 

\section{Model}
\label{sec:model}

As indicated by the related work on text normalization, character-level machine translation is a viable way of normalizing text into a standard variety. Therefore, we will also use character-level NMT in building our sequence-to-sequence OCR post-correction model. However, such a model requires parallel data for training. First, we will present our method of automatically extracting parallel data from our corpus containing OCR errors, then we will present the model designed to carry out the actual error correction.

\subsection{Extracting Parallel Data}
\label{sec:extract}
To extract a parallel corpus of OCR errors and their correctly spelled counterparts out of our corpus, we use a simple procedure consisting of measuring the similarity of the OCR errors with their correct spelling candidates. The similarity is measured in two ways, on the one hand an erroneous form will share a similarity in meaning with the correct spelling as they are realizations of the same word. On the other hand, an erroneous form is bound to share similarity on the level of characters, as noted by \citet{hill-hengchen2019OCR} in their study of OCR typically failing on a few characters on the corpus at hand.

In order to capture the semantic similarity, we use Gensim \cite{rehurek_lrec} to train a Word2Vec \cite{mikolov2013efficient} model.\footnote{Parameters: CBOW architecture, window size of 5, frequency threshold of 100, 5 epochs. Tokens were lowercased and no stopwords were removed.} As this model is trained on the corpus containing OCR errors, when queried for the most similar words with a correctly spelled word as input, the returned list is expected to contain OCR errors of the correctly spelled word together with real synonyms, the key finding which we will exploit for parallel data extraction.

As an example to illustrate the output of the Word2Vec model, a query with the word \textit{friendship} yields \textit{friendlhip, friendihip, friendflip, friend-, affection, friendthip, gratitude, affetion, friendflhip} and \textit{friendfiip} as the most similar words. In other words, in addition to the OCR errors of the word queried for, other correctly-spelled, semantically similar words (\textit{friend-, affection} and \textit{gratitude}) and even their erroneous forms (\textit{affetion}) are returned. Next, we will describe our method (as shown in Algorithm \ref{algo2}) to reduce noise in this initial set of parallel word forms.

As illustrated by the previous example, we need a way of telling correct and incorrect spellings apart. In addition, we will need to know which incorrect spelling corresponds to which correct spelling (\textit{affetion} should be grouped with \textit{affection} instead of \textit{friendship}).

For determining whether a word is a correctly spelled English word, we compare it to the lemmas of the Oxford English Dictionary (OED).\footnote{\url{http://www.oed.com}.} If the word exists in the OED, it is spelled correctly. However, as we are comparing to the OED lemmas, inflectional forms would be considered as errors, therefore, we lemmatize the word with spaCy\footnote{Using the \texttt{en\_core\_web\_md model}.} \cite{spacy2}. If neither the word nor its lemma appear in the OED, we consider it as an OCR error.

For a given correct spelling, we get the most similar words from the Word2Vec model. We then group these words into two categories: correct English words and OCR errors. For each OCR error, we group it with the most similar correct word on the list. This similarity is measured by using Levenshtein edit distance \citep{levenshtein1966binary}. The edit distances of the OCR errors to the correct words they were grouped with are then computed. If the distance is higher than 3 -- a simple heuristic, based on ad-hoc testing --, we remove the OCR error from the list. Finally, we have extracted a small set of parallel data of correct English words and their different erroneous forms produced by the OCR process.

\begin{algorithm}
\caption{Extraction of parallel data}
\label{algo2}
\SetAlgoLined

Draw $words$ $w$ from the input word list; \\
\For{$w$}{
    Draw synonyms $s_w$ in the word embedding model\\
	\For{synonym $s_w$}{
	    \If{$s_w$ is correctly spelled }{
	        Add $s_w$ to correct forms $forms_c$
	    }
	    \Else{
	    Add $s_w$ to error forms $forms_e$
	    }
	}
	\For{error $e$ in $forms_e$}{
	    group $e$ with the correct form in $forms_c$ by $Lev_{min}$
	    
		\If{Lev$_{(e,c)}$ $>$ 3}{
		    remove($e$)}
		 }
	
}
\end{algorithm}

We use the extraction algorithm to extract the parallel data by using several different word lists. First, we list all the words in the vocabulary of the Word2Vec model and list the words that are correctly spelled. We use this list of correctly spelled words in the model to do the extraction. However, as this list introduces noise to the parallel data, we combat this noise by producing another list of correctly spelled words that have occurred over 100,000 times in ECCO. For these two word lists, one containing all the correct words in the model and the other filtered with word frequencies, we produce parallel datasets consisting of words longer or equal to 1, 2, 3, 4 and 5. The idea behind these different datasets is that longer words are more likely to be matched correctly with their OCR error forms, and also frequent words will have more erroneous forms than less frequent ones.

In addition, we use the frequencies from the British National Corpus \cite{bnc} to produce one more dataset of words occurring in the BNC over 1000 times to test whether the results can be improved with frequencies obtained from a non-noisy corpus. This BNC dataset is also used to produce multiple datasets based on the length of the word. The sizes of these automatically extracted parallel datasets are shown in Table \ref{tab:size-table}.

\begin{table}[]
\centering
\small
\begin{tabular}{|l|l|l|l|l|l|}
\hline
source & all & \textgreater{}=2 & \textgreater{}=3 & \textgreater{}=4 & \textgreater{}=5 \\ \hline
W2V all & 29013 & 28910 & 27299 & 20732 & 12843 \\ \hline
\begin{tabular}[c]{@{}l@{}}W2V freq \\ \textgreater{}100,000\end{tabular} & 11730 & 11627 & 10373 & 7881 & 5758 \\ \hline
BNC & 7692 & 7491 & 6681 & 5926 & 4925 \\ \hline
\end{tabular}
\caption{Sizes of the extracted parallel datasets}
\label{tab:size-table}
\end{table}

\begin{table*}[]
\centering
\scriptsize
\setlength\tabcolsep{2pt}
\begin{tabular}{|l|l|l|l|l|l|l|l|l|l|l|l|l|l|l|l|}
\hline
\multicolumn{1}{|c|}{source} & \multicolumn{3}{c|}{all} & \multicolumn{3}{c|}{\textgreater{}=2} & \multicolumn{3}{c|}{\textgreater{}=3} & \multicolumn{3}{c|}{\textgreater{}=4} & \multicolumn{3}{c|}{\textgreater{}=5} \\ \hline
\multicolumn{1}{|c|}{} & \multicolumn{1}{c|}{Correct} & \multicolumn{1}{c|}{\begin{tabular}[c]{@{}c@{}}False \\ positive\end{tabular}} & \multicolumn{1}{c|}{\begin{tabular}[c]{@{}c@{}}No \\ output\end{tabular}} & \multicolumn{1}{c|}{Correct} & \multicolumn{1}{c|}{\begin{tabular}[c]{@{}c@{}}False \\ positive\end{tabular}} & \multicolumn{1}{c|}{\begin{tabular}[c]{@{}c@{}}No \\ output\end{tabular}} & \multicolumn{1}{c|}{Correct} & \multicolumn{1}{c|}{\begin{tabular}[c]{@{}c@{}}False \\ positive\end{tabular}} & \multicolumn{1}{c|}{\begin{tabular}[c]{@{}c@{}}No \\ output\end{tabular}} & \multicolumn{1}{c|}{Correct} & \multicolumn{1}{c|}{\begin{tabular}[c]{@{}c@{}}False \\ positive\end{tabular}} & \multicolumn{1}{c|}{\begin{tabular}[c]{@{}c@{}}No \\ output\end{tabular}} & \multicolumn{1}{c|}{Correct} & \multicolumn{1}{c|}{\begin{tabular}[c]{@{}c@{}}False \\ positive\end{tabular}} & \multicolumn{1}{c|}{\begin{tabular}[c]{@{}c@{}}No \\ output\end{tabular}} \\ \hline
W2V all & 0,510 & 0,350 & 0,140 & 0,500 & 0,375 & 0,125 & 0,520 & 0,325 & 0,155 & 0,490 & 0,390 & 0,120 & 0,525 & 0,390 & 0,085 \\ \hline
\begin{tabular}[c]{@{}l@{}}W2V freq \\ \textgreater{}100,000\end{tabular} & 0,515 & 0,305 & 0,180 & 0,540 & 0,310 & 0,150 & 0,510 & 0,340 & 0,150 & 0,540 & 0,315 & 0,145 & 0,515 & 0,330 & 0,155 \\ \hline
BNC & \textbf{0,580} & 0,285 & 0,135 & 0,555 & 0,300 & 0,145 & 0,570 & \textbf{0,245} & 0,185 & 0,550 & 0,310 & 0,140 & 0,550 & 0,315 & 0,135 \\ \hline
\end{tabular}
\caption{Results of the NMT models trained on different datasets}
\label{tab:results}
\end{table*}

\subsection{The NMT Model}

We use the automatically extracted parallel datasets to train a character level NMT model for each dataset. For this task, we use OpenNMT\footnote{Version 0.2.1 of opennmt-py} \cite{opennmt} with the default parameters except for the encoder where we use a BRNN (bi-directional recurrent neural network) instead of the default RNN (recurrent neural network) as BRNN has been shown to provide a performance gain in character-level text normalization \cite{ceec_normal2}. We use the default of two layers for both the encoder and the decoder and the default attention model, which is the general global attention presented by \citet{luong2015effective}. The models are trained for the default number of 100,000 training steps with the same seed value.

We use the trained models to do a character level translation on the erroneous words. We output the top 10 candidates produced by the model, go through them one by one and check whether the candidate word form is a correct English word (as explained in section \ref{sec:extract}). The first candidate that is also a correct English word is considered as the corrected form produced by the system. If none of the top 10 candidates is a word in English, we consider that the model failed to produce a corrected form. The use of looking at the top 10 candidates instead of the topmost candidates is motivated by the findings by \citet{ceec_normal2} in historical text normalization with a character-level NMT.

\section{Evaluation}
\label{sec:eval}

For evaluation, we prepare by hand a gold standard containing 200 words with OCR errors from the ECCO and their correct spelling. The performance of our models calculated as a percentage of how many erroneous words they were able to fix correctly. As opposed to the other common metrics such as character error rate and word error rate, we are measuring the absolute performance in predicting the correct word for a given erroneous input word.

Table \ref{tab:results} shows the results for each dataset. The highest accuracy of 58\% is achieved by training the model with all of the frequent words in the BNC, and the lowest number of false positives (i.e. words that do exist in English but are not the right correction for the OCR error) is achieved by the model trained with the BNC words that are at least 3 characters long. The \textit{No output} column shows the number of words the models didn't output any word for that would have been correct English.

If, instead of using NMT, we use the Word2Vec extraction method presented in section \ref{sec:extract} to conduct the error correction by finding the semantically similar word with the lowest edit distance under 4 for an erroneous form, the accuracy of such a method is only 26\%. This shows that training an NMT model is a meaningful part in the correction process.

In the spirit of \citet{ceec_normal_santafe}, whose results indicate that combining different methods in normalization can be beneficial, we can indeed get a minor boost for the results of the highest accuracy NMT model if we first try to correct with the above described Word2Vec method and then with NMT, we can increase the overall accuracy to 59.5\%. However, there is no increase if we invert the order and try to first correct with the NMT and after that with the Word2Vec model.

\section{Conclusion and Future Work}
\label{sec:conclusion}

In this paper we have proposed an unsupervised method for correcting OCR errors. Apart from the lemma list and the lemmatizer, which can also be replaced by a morphological FST (finite-state transducer) analyzer or a list of word forms, this method is not language specific and can be used even in scenarios with less NLP resources than what English has. Although not a requirement, having the additional information about word frequencies from another OCR error-free corpus can boost the results.

A limitation of our approach is that it cannot do word segmentation in the case where multiple words have been merged together as a result of the OCR process. However, this problem is complex enough on its own right to deserve an entire publication of its own and is thus not in the scope of our paper. Indeed, previous research has been conducted focusing solely on the segmentation problem \cite{nastase-hitschler-2018-correction,sonietal2019} of historical text and in the future such methods can be incorporated as a preprocessing step for our proposed method.

It is in the interest of the authors to extend the approach presented in this paper on historical data written in Finnish and in Swedish in the immediate near future. The source code and the best working NMT model discussed in this paper has be made freely available on GitHub as a part of the natas Python library\footnote{https://github.com/mikahama/natas}.

\section*{Acknowledgements}
We would like to thank the COMHIS group\footnote{\url{https://www.helsinki.fi/en/researchgroups/computational-history}} for their support, as well as GALE for providing the group with ECCO data.

\bibliography{bib}
\bibliographystyle{acl_natbib}

\end{document}